\documentclass[letterpaper, 10 pt, conference]{ieeeconf}  

\IEEEoverridecommandlockouts                              

\usepackage{graphicx} 
\usepackage{epsfig} 
\usepackage[table]{xcolor}
\definecolor{grey}{rgb}{0.5,0.5,0.5}
\usepackage{booktabs}
\usepackage{multirow}
\usepackage{subcaption}
\usepackage{hyperref}
\usepackage{amsmath}
\usepackage{algorithm}
\usepackage{bm}
\usepackage{algpseudocode}
\usepackage{float}
\usepackage{url}
\usepackage{caption}
\usepackage{array}
\usepackage{amssymb}  

\usepackage{amsthm}
\theoremstyle{plain}
\newtheorem{theorem}{Theorem}[section]
\newtheorem{proposition}[theorem]{Proposition}

\theoremstyle{definition}
\newtheorem{definition}[theorem]{Definition}
\newtheorem{assumption}[theorem]{Assumption}
\theoremstyle{remark}

\newcommand{\projectpage}{\href{https://sites.google.com/view/robo-striker/}{RoboStriker}}

\usepackage[
    style=ieee,
    natbib=true,
    citestyle=numeric-comp,
    doi=false,
    isbn=false,
    url=false]{biblatex} 

\title{\LARGE \bf
RoboStriker: Latent-Space Strategic Games for \\ Autonomous Humanoid Boxing
}

\author{Kangning Yin$^{1,2,3*}$, Kaige Liu$^{4,2,3*}$, Zhe Cao$^{2*}$, Wentao Dong$^{1, 2}$, Weishuai Zeng$^{2}$, Tianyi Zhang$^{1}$, Qiang Zhang$^{5}$, \\Jingbo Wang$^{2}$, Jiangmiao Pang$^{2}$, Yang Li$^{1,2\dagger}$, Ming Zhou$^{2\dagger}$, Weinan Zhang$^{1,2\dagger}$ \\
\\
$^{1}$Shanghai Jiao Tong University, $^{2}$Shanghai Artificial Intelligence Laboratory, $^{3}$Shanghai Innovation Institute, \\
$^{4}$Huazhong University of Science and Technology,
$^{5}$University of Science and Technology of China}

\begin{document}

\maketitle

\thispagestyle{empty}
\pagestyle{empty}

\begin{abstract}

Achieving human-level competitive intelligence and physical agility in humanoid robots remains a profound challenge, particularly in contact-rich and highly dynamic tasks such as boxing. While Multi-Agent Reinforcement Learning offers a principled framework for strategic interaction, its direct application to unstructured raw motor spaces inevitably leads to joint-level physical collapse, preventing the emergence of any viable combat tactics. To resolve this fundamental conflict between strategic exploration and physical feasibility, we formulate the humanoid combat task as a novel two-player latent-space zero-sum Markov game. Under standard regularity and approximate best-response assumptions, we show that the latent formulation induces an equivalent game over the decoder-reachable action manifold, providing an approximate-Nash interpretation of the resulting self-play dynamics. To instantiate this theoretical formulation, we propose RoboStriker, a hierarchical framework that decouples high-level reasoning from low-level execution. It first distills the tracking expertise of predefined boxing motions into a topologically bounded latent manifold. This structured latent foundation subsequently drives multi-agent co-evolution via Latent-Space Neural Fictitious Self-Play. Extensive experimental results demonstrate that gaming within this structured latent space substantially outperforms direct exploration. By constraining strategic exploration through a pretrained motion decoder, RoboStriker substantially reduces the catastrophic balance failures observed in raw action-space methods and achieves superior tactical performance in both competitive win rates and striking efficiency. Finally, we successfully deploy and validate our learned combat policies on real-world humanoid robots. Our code and video and supplementary materials are available at \href{https://sites.google.com/view/robo-striker/}{RoboStriker}.

\end{abstract}

%
\section{Introduction}
Humanoid robots serve as a critical frontier for embodied intelligence. Recent breakthroughs have demonstrated outstanding capabilities in locomotion and agile maneuvers \cite{liao2025beyondmimicmotiontrackingversatile, yin2025unitracker, zeng2025behaviorfoundationmodelhumanoid, zhang2025track}. These successes, however, rely heavily on static or predictable environments where the primary objective is single-agent whole-body motion tracking and trajectory generation~\cite{peng2018deepmimic,yin2025unitracker}. When extending humanoid control to competitive, multi-agent physical interactions, such as boxing, the problem fundamentally shifts from isolated trajectory tracking to strategic co-adaptation under severe physical constraints. This shift naturally frames humanoid competition as a multi-agent reinforcement learning problem.

Yet, enabling humanoid robots to engage in competitive physical interactions presents a profound scientific challenge: the fundamental conflict between unconstrained strategic interaction and strict physical feasibility. 
From a robotics perspective, a humanoid robot, such as the Unitree G1, operates within a high-dimensional, unstructured raw motor space. The core difficulty is that the vast majority of joint combinations in this unconstrained space correspond to physically invalid motions.
To discover competitive tactics, reinforcement learning inherently requires active, diverse exploration. 
However, executing such unconstrained exploration in a fragile action space inevitably leads to catastrophic balance failures long before any meaningful strategic interaction can materialize \cite{peng2018deepmimic, peng2021amp}. 
Consequently, robots are heavily penalized for these basic physical violations, creating an embodied cold-start problem where discovering even a single valid combat maneuver is overwhelmingly difficult \cite{luo2023universal}. This inherent physical fragility is then critically exacerbated in the multi-agent setting, where an active, evolving opponent introduces continuous, unpredictable physical perturbations \cite{samuel1959some,hernandez2019generalized}. Without any structural bounds in the raw motor space to physically withstand these impacts, the learning process frequently collapses under the weight of basic balance maintenance, completely preventing the emergence of higher-level tactical intelligence.

To resolve this bottleneck, we formulate humanoid boxing as a two-player latent-space zero-sum Markov game, where strategic competition occurs over a bounded latent domain and a pretrained low-level motion decoder maps latent actions to a decoder-reachable motor manifold. Under standard regularity and approximate best-response assumptions, the resulting latent game admits an approximate-Nash interpretation over this decoder-induced action manifold (details in Section~\ref{sec:two_player_zero_sum_games}).
Building on this theoretical formulation, we introduce RoboStriker, a hierarchical framework that decomposes embodied MARL into three coupled layers: a physically grounded motion library, a structured latent motion space for strategy representation, and multi-agent strategy evolution over this latent space. In the first layer, we establish a tracking policy to faithfully reproduce diverse human motion primitives. In the second layer, we distill these skills into a bounded latent motion space that supports diverse motion generation. Finally, we introduce Latent-Space NFSP (LS-NFSP), enabling multi-agent co-evolution within this compact manifold. Furthermore, an Adversarial Motion Prior (AMP)~\cite{peng2021amp} curriculum warmup initializes the policy, mitigating the competitive cold-start problem and substantially reducing non-stationarity.
We evaluate our framework on a humanoid boxing task using Unitree G1 robots~\cite{unitreeg1}. Experimental results demonstrate that by restricting strategic exploration to a physically grounded latent space, RoboStriker fundamentally avoids the balance collapse inherent in existing baselines. Our core contributions are summarized as follows:
\begin{itemize}
    \item We formulate competitive humanoid boxing as a two-player latent-space zero-sum Markov game induced by a pretrained motion decoder. Under standard regularity and approximate best-response assumptions, we characterize the correspondence between latent strategies and the decoder-reachable physical action manifold, providing an approximate-Nash analysis of the resulting game.
    \item We propose RoboStriker, a hierarchical algorithmic framework that instantiates this latent game by decoupling high-level strategic reasoning from low-level physical execution. By distilling motion tracking expertise into a topologically structured continuous manifold, the framework seamlessly drives robust multi-agent co-evolution via Latent-Space Neural Fictitious Self-Play.
    \item We empirically validate our framework on 29-degree-of-freedom Unitree G1 humanoids, demonstrating that RoboStriker achieves superior tactical performance by surpassing existing baselines in competitive win rates and striking efficiency. Our experiments further show that latent-space strategic learning substantially improves physical stability over raw action-space exploration, while the 29-DoF+AMP baseline confirms that these gains
    cannot be explained by motion regularization alone. Finally, we validate the learned policies on real-world humanoid hardware, as shown in ~\autoref{fig:real_world}.
\end{itemize}

\section{Related Work}

\subsection{Humanoid Motion Synthesis and Control}

Physics-based humanoid control has progressed from motion-specific imitation~\cite{peng2018deepmimic}
to adversarial motion priors~\cite{peng2021amp} and recent universal motion trackers~\cite{yin2025unitracker,zhang2025track,chen2025gmt}.
These methods enable robust reproduction of diverse human motions under physical constraints, providing an effective foundation for reusable whole-body control.
However, they primarily focus on single-agent motion tracking or task execution with externally specified motion objectives.
Competitive humanoid interaction introduces an additional challenge: the controller must not only preserve balance and motion quality, but also continuously adapt its behavior to a non-stationary and physically interacting opponent.
RoboStriker builds on robust motion tracking as a low-level motor foundation and studies how such motor expertise can be exposed as a compact interface for multi-agent strategic learning.

\subsection{Latent Motion Representations and Motion Priors}

Learning reusable latent motion representations is a common approach for reducing the complexity of high-dimensional character and humanoid control.
ASE~\cite{peng2022ase} learns reusable adversarial skill embeddings with a hyperspherical latent representation, while CALM~\cite{tessler2023calm} develops conditional latent models for controllable character behaviors.
PULSE~\cite{luo2023universal} instead learns a state-conditioned latent motion representation that distills broad humanoid motion expertise into a compact interface for downstream control.
These works demonstrate that learned latent variables can provide effective abstractions over complex motor behaviors.

Our work does not rely on the hyperspherical parameterization itself as a new motion-representation mechanism.
Rather, we use a bounded latent motion interface as the strategic action space of a competitive humanoid game.
The pretrained state-conditioned decoder maps high-level latent decisions to a decoder-reachable motor manifold, while multi-agent learning is performed entirely over this compact interface.
This formulation allows us to explicitly study how latent action abstraction interacts with adversarial motion regularization and competitive self-play.

\subsection{Competitive Character Control and Multi-Agent Reinforcement Learning}
\vspace{1.5mm}
Competitive control has been extensively studied in multi-agent reinforcement learning and physically simulated character animation.
Classical self-play~\cite{samuel1959some,hernandez2019generalized} enables agents to improve through repeated interaction, while fictitious play~\cite{brown1951iterative} and Neural Fictitious Self-Play~\cite{heinrich2016deep} stabilize learning by incorporating historical or averaged opponent strategies.
Prior work has also demonstrated physically simulated characters performing competitive sports, including two-player athletic interactions~\cite{10.1145/3450626.3459761}, and more recent humanoid systems have explored learned athletic skills in interactive sports settings~\cite{zhang2026learningathletichumanoidtennis}.

RoboStriker builds on these studies by performing competitive learning over a pretrained latent motion interface rather than the full motor space. Combined with Neural Fictitious Self-Play, this decouples opponent-dependent strategy learning from low-level whole-body execution while restricting optimization to a compact latent domain. This formulation also provides the basis for our approximate-Nash analysis of the induced latent game.

\begin{figure*}
    \centering
    \includegraphics[width=\textwidth]{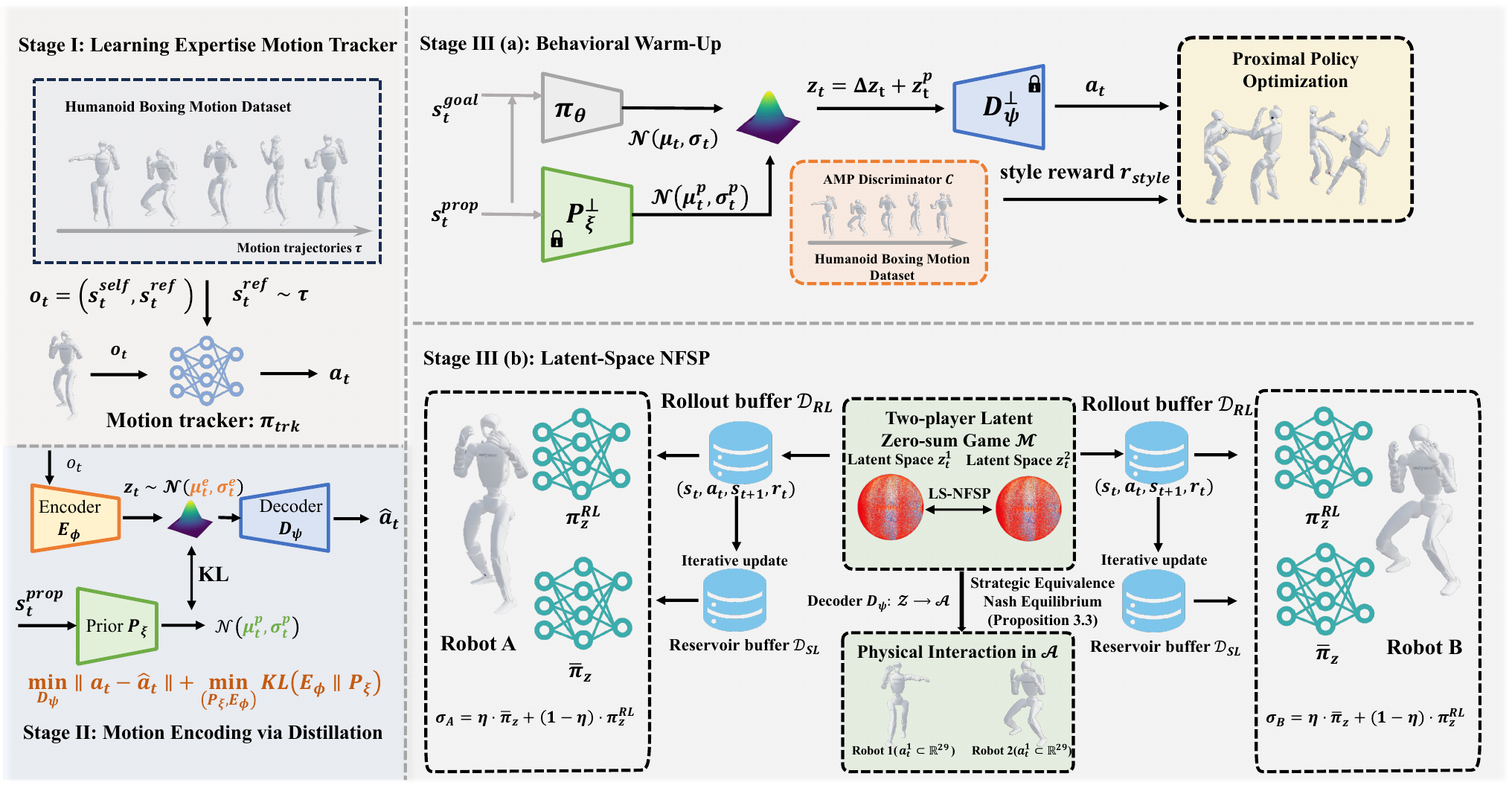 }
    \caption{Overview of RoboStriker. \textbf{Stage I} pretrains a motion tracker to produce physically plausible humanoid behaviors; \textbf{Stage II} compresses these behaviors into a bounded latent space for high-level control; \textbf{Stage III(a)} runs warm-up training on top of Stage II, then followed with \textbf{Stage III(b)}, a NFSP over latent-space to solve the humanoid boxing as a two-player latent space zero-sum game.}
    \label{fig:pipeline}
    \vspace{-4mm}
\end{figure*}

\section{Method}
Achieving autonomous humanoid boxing requires overcoming the fundamental conflict between unconstrained multi-agent strategic exploration and the extreme physical fragility of high-dimensional motor control. To resolve this inherent instability, we propose RoboStriker, a hierarchical competitive learning framework that integrates theoretical game formulation with a robust execution pipeline. We first formalize the combat task as a two-player latent-space zero-sum Markov game in Section~\ref{sec:two_player_zero_sum_games}, where strategic interaction is restricted to a bounded latent domain induced by the pretrained motion decoder. A formal approximate-Nash analysis of this induced game is provided on our project page. To instantiate this formulation, our pipeline systematically decouples high-level intent from low-level execution through three seamlessly coupled stages: learning a robust physical foundation via a universal motion tracker detailed in Section~\ref{sec:stage1}; distilling diverse motor expertise into the topologically structured latent space outlined in Section~\ref{sec:stage2}; and progressively evolving multi-agent strategies via an AMP-regularized warmup and Latent-Space Neural Fictitious Self-Play elaborated in Section~\ref{sec:stage3a} and~\ref{sec:stage3b}.

\subsection{Two-player Latent Space Zero-sum Markov Games}
\label{sec:two_player_zero_sum_games}
We formulate the two-player humanoid boxing task as a latent-space zero-sum Markov Game, defined by tuple $\mathcal{M}=\langle \mathcal{I}, \mathcal{S}, \{\mathcal{O}\}^{i \in \mathcal{I}}, \{\mathcal{A}\}^{i \in \mathcal{I}}, \{\mathcal{Z}\}^{i \in \mathcal{I}}, D_\psi, \mathcal{P}, \{\mathcal{U}\}^{i\in \mathcal{I}}, \gamma \rangle$. $\mathcal{I}=\{1,2\}$ is the set of players, $\mathcal{S}$ is the underlying state space, and $\{\mathcal{O}\}^{i \in \mathcal{I}}$ is the set of player observation spaces.

Crucially, to distinguish high-level strategic reasoning from low-level physical execution, we explicitly define $\{\mathcal{A}\}^{i \in \mathcal{I}}$ as the set of raw motor action spaces, specially designating $\mathcal{A}^i \subset \mathbb{R}^{29}$ as the 29-degree-of-freedom joint position targets. We also define $\{\mathcal{Z}\}^{i \in \mathcal{I}}$ as the set of bounded continuous latent strategy spaces, where $\mathcal{Z}^i \subset \mathbb{R}^d$. Actions executed in the environment are mapped from the latent space via a pre-trained, shared motion decoder $D_\psi: \mathcal{S} \times \mathcal{Z}^i \rightarrow \mathcal{A}^i$. Consequently, the environment's underlying transition dynamics $\mathcal{P}_{env}: \mathcal{S} \times \mathcal{A}^1\times\cdots\times\mathcal{A}^{\vert \mathcal{I} \vert} \rightarrow \Delta(\mathcal{S})$ and raw utility functions $u^i_{env}: \mathcal{S} \times \mathcal{A}^1 \times \cdots \times \mathcal{A}^{\vert \mathcal{I} \vert} \rightarrow \mathbb{R}$ induce a latent-level transition function $\mathcal{P}: \mathcal{S} \times \mathcal{Z}^1\times\cdots\times\mathcal{Z}^{\vert \mathcal{I} \vert} \rightarrow \Delta(\mathcal{S})$ and utility function $u^i \in \mathcal{U}^i: \mathcal{S} \times \mathcal{Z}^1 \times \cdots \times \mathcal{Z}^{\vert \mathcal{I} \vert} \rightarrow \mathbb{R}$. Here $\Delta(\cdot)$ denotes the probability simplex over a given set. Specifically, the induced utility is given by $u^i(s, z^1, z^2) = u^i_{env}(s, D_\psi(s, z^1), D_\psi(s, z^2))$, with $\gamma \in [0, 1)$ as the discount factor.

Let each player $i\in \mathcal{I}$ adopt a latent policy $\pi^i_z: \mathcal{O}^i \rightarrow \Delta(\mathcal{Z}^i)$, with the joint policy denoted by $\boldsymbol{\pi}_z=(\pi^1_z,\pi^2_z)$. The expected return for player $i$ under joint policy $\boldsymbol{\pi}_z$ from state $s \in \mathcal{S}$ is given by $V^i_{\boldsymbol{\pi}_z}(s) = \mathbb{E}\left[\sum_{t=0}^{\infty}\gamma^t\,u^i(s_t, z^i_t,z^{-i}_t)\mid s_0=s\right]$, where $-i$ indicates the opponent. In the strictly zero-sum setting $u^1=-u^2$, rather than directly solving a static optimization objective, each agent iteratively updates its policy by computing an approximate best response to the empirical strategy distribution of its opponent. If the policy updates are treated as bounded-error approximate best responses to the empirical opponent strategy, the resulting empirical policies admit an $\epsilon$-Nash characterization in the latent game. The corresponding approximation error depends on the quality of the learned best responses and regularization introduced during optimization.

A formal analysis of latent-space approximate Nash convergence, including assumptions and a proof sketch, is provided on project page. We next instantiate the latent-game formulation through a three-stage hierarchical pipeline.
\subsection{Learning Expertise Motion Tracker}
\label{sec:stage1}
In the first stage shown in~\autoref{fig:pipeline} (Stage I), we learn a robust low-level controller that can faithfully track diverse humanoid boxing motions and provide a stable motor foundation for high-level strategic learning.
The tracker $\pi_{\mathrm{trk}}(a_t \mid s_t^{\mathrm{self}}, s^{\mathrm{ref}}_t)$ is trained using human motion capture data and shared across all robots,
where $s_t^{\mathrm{self}}=(s_t^{\mathrm{prop}}, s_t^{\mathrm{priv}})$. Here, $s_t^{\mathrm{prop}}$ denotes a robot's proprioceptive state at timestep $t$ and $s_t^{\mathrm{priv}}$ represents privileged observations that are inaccessible to the robot's onboard sensors. $s^{\mathrm{ref}}_t$ specifies a reference motion goal derived from human motion capture data, which is represented as a time-indexed sequence of full-body humanoid poses and velocities extracted from motion capture dataset $\mathcal{D}_{\mathrm{motion}}$.
Each $s^{\mathrm{ref}}_t$ encodes the root position and orientation, joint angles, and corresponding joint velocities in a canonical humanoid kinematic tree.
All motions are temporally aligned and retargeted to the simulated humanoid morphology, enabling consistent tracking across diverse boxing behaviors.
The objective maximizes expected tracking rewards $r_{\mathrm{trk}}$ over reference trajectories $\tau^{\mathrm{ref}}$, encouraging high-fidelity imitation while maintaining physical plausibility,
\begin{align}
    &\pi_{\mathrm{trk}}^{*} = \arg\max_{\pi_{\mathrm{trk}}} \mathbb{E}_{s^{\mathrm{ref}} \sim \tau^{\mathrm{ref}}, \tau^{\mathrm{ref}}\sim \mathcal{D}_{\mathrm{motion}}} \left[ J_{\pi_{\mathrm{trk}}}(s^{\mathrm{ref}})\right],\\\nonumber
    &\text{where } J_{\pi_{\mathrm{trk}}}(s^{\mathrm{ref}})=\mathbb{E}_{a_t \sim \pi_{\mathrm{trk}}} \left[ \sum_{t=0}^{T} \gamma^{t} \, r_{\mathrm{trk}}(s_t^{\mathrm{self}}, a_t; s^{\mathrm{ref}}) \right].
\end{align}

Concretely, we collect boxing motions from professional boxers with an Xsens motion capture system~\cite{xsens_mvn} and double the corpus via left--right mirroring, yielding $\sim$30 minutes at 50\,Hz motion sequences, covering strikes, defense, footwork, and transitions. Captured sequences are retargeted to the Unitree G1 morphology with Generalized Motion Retargeting~\cite{araujo2025retargetingmattersgeneralmotion} to form $\mathcal{D}_{\mathrm{motion}}$. The tracking observation concatenates proprioception $s_t^{\mathrm{prop}}$, a short-horizon reference window of $K{=}3$ future poses, and the previous action $a_{t-1}$; the action is a 29-DoF joint-position target tracked by a PD controller. The reward $r_{\mathrm{trk}}$ is a weighted sum of exponential terms for root and body pose/orientation tracking, linear/angular velocity matching, and control regularization (action rate, joint limits, and undesired contacts). Following~\cite{peng2018deepmimic}, episodes terminate early under pose collapse, end-effector height violations, or a 10\,s horizon cap. Exact reward weights and scaling coefficients are provided on our project page.

\subsection{Encoding Motion via Topological Latent Distillation}
\label{sec:stage2}
As illustrated in~\autoref{fig:pipeline} (Stage II), to enable learnable strategic control, we project the high-dimensional motion space into a compact, continuous latent space $\mathcal{Z}$.
This is achieved via a teacher-student distillation framework~\cite{ross2011reductionimitationlearningstructured} consisting of an encoder $E_\phi$, a decoder $D_\psi$, and a state-conditioned latent prior $P_\xi$.
The encoder maps observations $o_t=(s_t^{\mathrm{self}}, s^{\mathrm{ref}}_t)$ to a distribution of latent codes $z_t \sim E_{\phi}(\cdot \mid o_t)$.
In implementation, we model $E_{\phi}(z_t)$ as diagonal Gaussian $\mathcal{N}(z_t \mid \mu^e_t, \sigma^e_t)$. Then the decoder is trained to reconstruct the teacher’s actions $a_t \sim \pi_{\mathrm{trk}}(\cdot \mid o_t)$ conditioned on $(s^{\mathrm{prop}}_t, z_t)$ as
\begin{equation}
    D^{\star}_{\psi} = \arg\min_{D_{\psi}} \parallel a_t - \hat{a}_t\parallel,
\end{equation}
where $\hat{a}_t \sim D_{\psi}(\cdot \mid s^{\mathrm{prop}}_t,z_t)$.
Acknowledging the inherent state-dependency of motion generation, we explicitly learn a state-conditioned latent prior, denoted as $P_{\xi}(z_t\mid s^{\mathrm{prop}}_t)=\mathcal{N}(z_t \mid \mu^p_t,\sigma^p_t)$. This prior serves to model valid transitions from the current proprioceptive state and constrains the encoder $E_\phi(z_t \mid o_t)$ via KL-regularization~\cite{hershey2007approximating}, thereby preventing posterior collapse, i.e.,
\begin{equation}
    \label{eq:learning_prior}
    (E^{\star}_{\phi}, P^{\star}_\xi) = \arg\min_{(E_\phi, P_\xi)} D_{KL}\left(E_\phi(z_t \mid o_t) \parallel P_\xi(z_t \mid s^{\mathrm{prop}}_t)\right).
\end{equation}

A key design is to bound the strategic latent domain. We therefore normalize latent codes onto the unit hypersphere,
providing a compact parameterization for high-level optimization. The pretrained state-conditioned decoder subsequently maps these latent decisions onto its learned motor manifold, which serves as the effective physical action space of the induced game. In implementation, we let $z$ satisfy the requirements with normalization as
\begin{equation}
    \hat{z} = \mathrm{Normalize}(z)=\frac{E_\phi(o)}{\| E_\phi(o) \|_2},
\end{equation}
thereby constraining the latent representation of $z$ to lie on the unit hypersphere. Jointly, the student optimizes action reconstruction and prior alignment,
\begin{align}
\mathcal{L}_{\mathrm{rec}} &= \mathbb{E}\!\left[\| a_t - \hat{a}_t\|^2\right],\\
\mathcal{L}_{\mathrm{prior}} &= \mathbb{E}\!\left[D_{\mathrm{KL}}\big(E_\phi(z_t\mid o_t)\,\|\,P_\xi(z_t\mid s_t^{\mathrm{prop}})\big)\right],\\
\mathcal{L}_{\mathrm{distill}} &= \mathcal{L}_{\mathrm{rec}} + \lambda_{\mathrm{prior}} \mathcal{L}_{\mathrm{prior}},
\end{align}
with $\lambda_{\mathrm{prior}}=0.001$. This dual objective recovers the teacher's motor expertise while constructing a structured, prior-conditioned manifold for subsequent competitive co-evolution.
\subsection{Behavioral Warmup with Adversarial Priors}
\label{sec:stage3a}
To bypass the severe instability caused by initiating self-play without basic tactical competence, we design a behavioral warmup stage, as depicted in~\autoref{fig:pipeline} (Stage IIIa). The agents learn effective striking behaviors against a stationary opponent that strictly maintains a stable standing stance. A learnable residual policy $\pi_{\theta}(\cdot \mid s^{\mathrm{goal}}_t)$ outputs residual latent commands $\Delta z_t$ over a fixed behavioral prior $P^{\perp}_{\xi}$ from the second stage, ensuring stable and human-like motions.
Then, the complete form of $\pi_{z}$ is
\begin{equation}
    \pi_z(\cdot \mid s^{\mathrm{prop}}, s^{\mathrm{goal}}) = \pi_{\theta}(\cdot \mid s^{\mathrm{goal}}) \oplus P^{\perp}_{\xi}(\cdot \mid s^{\mathrm{prop}}_t),
\end{equation}
where the operator $\oplus$ defines a normalized residual addition in the continuous action space. The corresponding generation of $z_t$ is formulated as
\begin{equation}
    z_t=\mathrm{Normalize}(\Delta z_t + z^p_t),
\end{equation}
where $\Delta z_t \sim \pi_{\theta}(\cdot \mid s^{\mathrm{goal}})$ and $z^p_t \sim P^{\perp}_\xi(\cdot \mid s^{\mathrm{prop}})$.

Unlike tracking, autonomous boxing provides no explicit motion commands. Instead, the residual policy conditions on a goal observation $s^{\mathrm{goal}}$ that concatenates ego-centric offensive and defensive geometry: relative vectors from the ego fists to the opponent torso, and from the opponent fists to the ego torso, all rotated into the ego root frame so that striking and threat cues are orientation-invariant.
To prevent degradation of motion quality, we regularize the warmup with AMP~\cite{peng2021amp}. A discriminator $C$ distinguishes motion transitions produced by $(\pi_z,D_\psi)$ from reference clips in $\mathcal{D}_{\mathrm{motion}}$, using discriminator features that include local joint rotations/velocities and base angular velocity. With a gradient penalty on real transitions, the discriminator is trained as
\begin{equation}
\begin{aligned}
\arg\min_C\;&
\mathbb{E}_{d^{\mathcal{D}_{\mathrm{motion}}}}\!\big[(C(o_t^{\mathrm{disc}},o_{t+1}^{\mathrm{disc}})-1)^2\big]
\\ &+
\mathbb{E}_{d^{\pi_z,D_\psi}}\!\big[(C(o_t^{\mathrm{disc}},o_{t+1}^{\mathrm{disc}})+1)^2\big] \\
&+ w_{gp}\,\mathbb{E}_{d^{\mathcal{D}_{\mathrm{motion}}}}\!\big[\|\nabla C(o_t^{\mathrm{disc}},o_{t+1}^{\mathrm{disc}})\|^2\big],
\end{aligned}
\end{equation}
and supplies a style reward $r_{\mathrm{style}}$ to the policy.
Thus, the learning objective is to maximize the expected return:
\begin{equation}
    \pi_z^\star = \arg\max_{\pi_{z}} \mathbb{E}_{\tau \sim (\pi_z, D_\psi)} \left[ \sum_{t=0}^{T} \gamma^t R(s_t,z_t) \right],
\end{equation}
where $R(s_t,z_t)=w_{\mathrm{task}} \cdot r_{\mathrm{task}}(s_t) + w_{\mathrm{style}} \cdot r_{\mathrm{style}}(s_t, s_{t+1})$, and the task term further decomposes into facing alignment, approach velocity, gated striking distance, and contact-validated hit rewards. Detailed warmup shaping formulas and numeric weights are provided on our project page.

\subsection{Latent-Space Neural Fictitious Self-Play}
\label{sec:stage3b}
Building upon the behavioral warmup stage, we adopt Latent-Space Neural Fictitious Self-Play to drive competitive co-evolution, as illustrated in~\autoref{fig:pipeline}(StageIIIb). Standard fictitious self-play in unconstrained motor spaces often collapses due to the severe physical fragility of humanoids. By implementing the competitive dynamics strictly over the structured latent action space $\mathcal{Z}$ rather than the raw motor space, our approach successfully neutralizes these instabilities. 

In our framework, each player is controlled by an independent agent that maintains a dual-policy system learning through simultaneous self-play interactions. An agent records its experience into two distinct buffers: a reinforcement learning dataset $\mathcal{D}_{RL}$ storing recent transitions, and a supervised learning dataset $\mathcal{D}_{SL}$ functioning as a reservoir of its historical best-response behaviors. For each agent, we train a best-response policy $\pi^{RL}_{z}$ via Proximal Policy Optimization~\cite{schulman2017proximalpolicyoptimizationalgorithms} using samples from $\mathcal{D}_{RL}$. Crucially, because the latent manifold $\mathcal{Z}$ is topologically bounded on a unit hypersphere, this reinforcement learning optimization inherently avoids the out-of-distribution physical actions that typically cause catastrophic forgetting and balance failures in continuous control.

In parallel, we train an average policy $\bar{\pi}_{z}$ from $\mathcal{D}_{SL}$ via supervised learning to imitate the historical best-response distribution. Over learning iterations $k = 1, \dots, K$, the supervised dataset is incrementally updated as
\begin{equation}
    \mathcal{D}^k_{SL} = \mathcal{D}^{k-1}_{SL} \cup \mathcal{D}^k_{RL},\forall k = 1,\dots,K.
\end{equation}
To keep $\bar{\pi}_{z}$ representative of the full behavioral history rather than only recent iterations, $\mathcal{D}_{SL}$ is maintained by reservoir sampling with capacity $K$: a new experience $n>K$ replaces a uniformly chosen entry with probability $K/n$. Action selection over the continuous latent space follows a mixture strategy controlled by an anticipatory parameter $\eta \in [0, 1)$. Formally, the mixed strategy is defined as
\begin{equation}
    \sigma=\eta \cdot \bar{\pi}_z + (1 - \eta) \cdot \pi^{RL}_z,
\end{equation}
with latent actions sampled as $z_t \sim \sigma(\cdot \mid s^{\mathrm{prop}}_t, s^{\mathrm{goal}}_t)$. Under this formulation, $\pi^{RL}_z$ represents an approximate best response to the mixed strategies of the opponent, while $\bar{\pi}_z$ acts as a stable strategic anchor. As established in our earlier theoretical analysis, the bounded latent parameterization provides the compact strategy domain required by our analysis. When the PPO updates approximate best responses with bounded error, the resulting empirical strategy can be interpreted as an approximate equilibrium of the decoder-induced latent game. Algorithm~\ref{alg:ls_nfsp} summarizes the complete training loop. Competitive-stage reward terms and their weights are detailed on our project page.

\begin{algorithm}[t]
\caption{Latent-Space Neural Fictitious Self-Play (LS-NFSP)}
\label{alg:ls_nfsp}
\begin{algorithmic}[1]
\Require Two-player zero-sum game environment $\mathcal{M}$, anticipatory parameter $\eta\in[0,1]$, horizon $T$
\Require Best-response (RL) policies $\{\pi^{\mathrm{RL}}_{z,\theta_i}\}_{i\in\{1,2\}}$, average policies $\{\bar{\pi}_{z,\phi_i}\}_{i\in\{1,2\}}$
\Require On-policy buffers $\{\mathcal{D}^{i}_{\mathrm{RL}}\}_{i\in\{1,2\}}$, reservoir buffers $\{\mathcal{D}^{i}_{\mathrm{SL}}\}_{i\in\{1,2\}}$
\For{iteration $k=1,2,\dots$}
    \State Reset $\mathcal{M}$ and observe $o_1^1,o_1^2$
    \For{$t=1,2,\dots,T$}
        \For{each player $i\in\{1,2\}$}
            \State Sample mode $m_i \sim \mathrm{Bernoulli}(\eta)$
            \If{$m_i = 1$}
                \State Sample $z_t^i \sim \pi^{\mathrm{RL}}_{z,\theta_i}(\cdot\mid o_t^i)$
                \State Insert $(o_t^i, z_t^i)$ into $\mathcal{D}^{i}_{\mathrm{SL}}$ via reservoir sampling
            \Else
                \State Sample $z_t^i \sim \bar{\pi}_{z,\phi_i}(\cdot\mid o_t^i)$
            \EndIf
        \EndFor
        \State Step $\mathcal{M}$ with $(z_t^1,z_t^2)$; receive $(r_t^1,r_t^2)$ and $(o_{t+1}^1,o_{t+1}^2)$
        \For{each player $i\in\{1,2\}$}
            \State Append $(o_t^i,z_t^i,r_t^i,o_{t+1}^i)$ to $\mathcal{D}^{i}_{\mathrm{RL}}$
        \EndFor
    \EndFor
    \For{each player $i\in\{1,2\}$}
        \State Update $\theta_i$ with PPO on $\mathcal{D}^{i}_{\mathrm{RL}}$
        \State Update $\phi_i$ by minimizing $\mathcal{L}_{\mathrm{SL}}=\mathbb{E}_{(o,z)\sim\mathcal{D}^{i}_{\mathrm{SL}}}\big[\|\bar{\pi}_{z,\phi_i}(o)-z\|^2\big]$
    \EndFor
\EndFor
\State \Return average policies $\{\bar{\pi}_{z,\phi_i}\}_{i\in\{1,2\}}$
\end{algorithmic}
\end{algorithm}

\begin{table*}[t]
  \centering
  \caption{Win Rate (\%) of Cross-Playing and Tactical Performance. Each Win Rate is the win rate of the row agent against the column agent, averaged over 20 evaluation rounds with 4,096 parallel bouts per round. Offensive Landing Rate ($\eta_{hit}$) and Engagement Rate (ER) are evaluated against Naive Self-Play (Latent). Bold values indicate the top-1 values in the table.}
  \label{tab:merged_results}
  \setlength{\tabcolsep}{2.5pt}
  \renewcommand{\arraystretch}{1.15}
  \begin{tabular}{l|ccccccccc|cc}
  \toprule
   \multirow{2}{*}{\textbf{Methods}} & \multicolumn{9}{c|}{\textbf{Win Rate (\%) vs.\ Opponent}} & \multirow{2}{*}{\bm{$\eta_{hit}$} $\uparrow$} & \multirow{2}{*}{\textbf{ER} $\uparrow$} \\
  & \textbf{(1)} & \textbf{(2)} & \textbf{(3)} & \textbf{(4)} & \textbf{(5)} & \textbf{(6)} & \textbf{(7)} & \textbf{(8)} & \textbf{(9)} & & \\
  \midrule
  (1) RoboStriker      & -- & \textbf{68.52} & \textbf{76.24} & \textbf{82.41} & \textbf{84.47} & \textbf{95.38} & \textbf{98.50} & \textbf{100.0} & \textbf{92.50} & \textbf{0.685 $\pm$ 0.03} & \textbf{0.824 $\pm$ 0.02} \\
  (2) Fictitious SP         & -- & -- & 62.35 & 71.84 & 75.16 & 92.11 & 96.18 & 99.42 & 88.14 & 0.420 $\pm$ 0.03 & 0.650 $\pm$ 0.04 \\
  (3) Naive SP              & -- & -- & -- & 68.45 & 70.38 & 88.61 & 94.52 & 98.53 & 85.25 & 0.350 $\pm$ 0.04 & 0.580 $\pm$ 0.05 \\
  (4) RoboStriker w/o AMP   & -- & -- & -- & -- & 62.12 & 82.14 & 88.37 & 95.18 & 78.40 & 0.490 $\pm$ 0.03 & 0.720 $\pm$ 0.05 \\
  (5) PPO-Only              & -- & -- & -- & -- & -- & 45.24 & 77.86 & 82.11 & 68.55 & 0.231 $\pm$ 0.03 & 0.495 $\pm$ 0.02 \\
  (6) Static-Target Specialist & -- & -- & -- & -- & -- & -- & 60.33 & 85.58 & 62.18 & 0.210 $\pm$ 0.04 & 0.450 $\pm$ 0.06 \\
  (7) RoboStriker w/o Warmup   & -- & -- & -- & -- & -- & -- & -- & 94.17 & 51.34 & 0.050 $\pm$ 0.02 & 0.120 $\pm$ 0.05 \\
  (8) 29Dof Action-Space SP & -- & -- & -- & -- & -- & -- & -- & -- & 11.50 & 0.142 $\pm$ 0.05 & 0.315 $\pm$ 0.08 \\
  (9) 29Dof Action-Space + AMP & -- & -- & -- & -- & -- & -- & -- & -- & -- & 0.285 $\pm$ 0.04 & 0.465 $\pm$ 0.05 \\
  \bottomrule
  \end{tabular}
  \vspace{-4mm}
\end{table*}

\section{Experiment}

\subsection{Experimental Setup}
\label{sec:exp_setup}
We train the 29-degree-of-freedom Unitree G1 humanoid~\cite{unitreeg1} in Isaac Lab~\cite{nvidia2025isaaclabgpuacceleratedsimulation} on a single NVIDIA RTX 4090 GPU, with 4{,}096 parallel environments, a 200\,Hz physics loop, and a 50\,Hz control policy. Strategic co-evolution operates over a 32-dimensional latent manifold. In the final LS-NFSP stage, we set the mixing coefficient to $\eta=0.1$ and maintain a strategy reservoir of capacity $K=10^{6}$ samples per agent, with task and AMP style weights $w_{\mathrm{task}}=0.8$ and $w_{\mathrm{style}}=0.2$, and prior coefficient $\lambda_{\mathrm{prior}}=0.001$.

To promote sim-to-real transfer, we randomize contact materials, joint biases, torso center-of-mass, and intermittent external pushes throughout training. The full randomization ranges are provided on our project page.

We compare RoboStriker against baselines that isolate opponent modeling and policy averaging, curriculum design , latent-space abstraction, and motion regularization.
\textbf{Naive SP} uses the same latent action space as ours but always trains against the opponent's latest policy~\cite{samuel1959some,hernandez2019generalized}, while
\textbf{Fictitious SP} samples uniformly from historical opponent policies~\cite{brown1951iterative} without learning an explicit average-policy network.
\textbf{PPO-Only} trains against a fixed opponent using PPO~\cite{schulman2017proximalpolicyoptimizationalgorithms}, and
\textbf{Static-Target Specialist} directly evaluates the warmup policy before competitive self-play.
\textbf{SP w/o Warmup} initiates self-play from scratch to test the role of the behavioral curriculum.

To disentangle latent action abstraction from motion-prior regularization, we further compare four controlled variants.
\textbf{RoboStriker w/o AMP} removes the AMP style reward~\cite{peng2021amp} while retaining the latent hierarchy.
\textbf{29-DoF Action-Space SP} performs self-play directly in the 29-DoF joint-position target space without the learned decoder.
\textbf{29-DoF Action-Space + AMP} augments this raw action-space baseline with the same AMP style objective used by RoboStriker, while still predicting joint targets directly.
Together with the full RoboStriker model, these variants separate whether performance gains arise from AMP-based behavioral regularization or from structuring strategic exploration through the learned latent motion manifold.

\subsection{Evaluation Metrics}
\label{sec:metrics}
We evaluate all methods across three complementary dimensions, motivated by principles from professional boxing and robust robotic control:

\textbf{Tactical Proficiency.}
\textit{Win Rate} measures overall superiority as the percentage of evaluation bouts won, where a bout terminates when any non-foot body part contacts the ground. \textit{Offensive Landing Rate}, denoted as $\eta_{\mathrm{hit}}$, is the proportion of offensive attempts whose contact force exceeds $F_{\mathrm{th}}=10\,\mathrm{N}$. \textit{Engagement Rate} ($ER$) is the temporal fraction during which the agent simultaneously maintains an effective striking distance $d_t\in[0.5, 1.2]\,\mathrm{m}$ and a facing alignment $\cos(\theta_t)>\tau_{\mathrm{face}}=0.9$, thereby penalizing tactical avoidance.

\textbf{Physical Stability.}
Physical stability is assessed via \textit{Base Orientation Stability} ($BOS$) and \textit{Torque Smoothness} ($TS_{\tau}$). $BOS$ measures upright posture under perturbations via the angular deviation of gravity vectors: $BOS = \mathbb{E} [ \exp\{-\| \mathbf{g}_{\mathrm{base}} - \mathbf{g}_{\mathrm{world}} \|^2\} ]$, higher is better. $TS_{\tau}$ quantifies high-frequency motor oscillations as the mean change in consecutive torques, $TS_{\tau} = \mathbb{E} [ \| \tau_{t} - \tau_{t-1} \| ]$, lower is better.

\begin{figure*}
        \centering
        \includegraphics[width=1.0\linewidth]{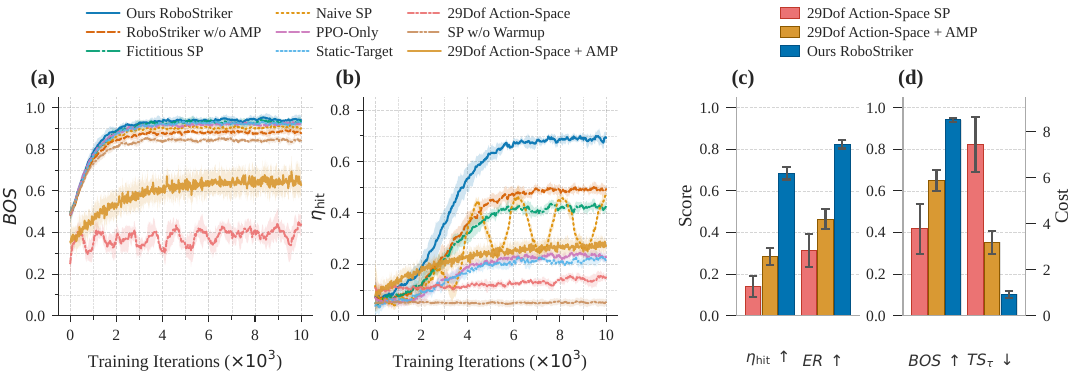}
        \caption{Comprehensive performance evaluation. Subfigures (a) and (b) show the training convergence of base stability and offensive landing rate across all methods. Subfigures (c) and (d) compare RoboStriker with direct 29-DoF action-space self-play with and without AMP, isolating the effects of motion-prior regularization and latent-space action abstraction on tactical proficiency and physical stability. Shaded regions denote the standard deviation across five independent random seeds.}
        \label{fig:training_convergence}
        \vspace{-2mm}
\end{figure*}

\textbf{Stylistic Authenticity.}
We qualitatively evaluate human-likeness through simulation snapshots, verifying the presence of key boxing maneuvers such as slips, counters, and rhythmic footwork, and the absence of unnatural or physically unsustainable postures typical of unconstrained reinforcement learning policies.

\subsection{Strategic Proficiency Comparison}
\label{sec:strategic_comparison}
Building on the baseline in~\autoref{sec:exp_setup}, we first examine strategic proficiency. As presented in~\autoref{tab:merged_results}, RoboStriker consistently outperforms competing methods, achieving win rates of $68.52\%$ against Fictitious SP, $76.24\%$ against Naive SP, and $84.47\%$ against PPO-Only. Beyond cross-play win rates, this strategic superiority is further evidenced by the quantitative tactical metrics, where RoboStriker achieves the highest Engagement Rate of $0.824$ and Offensive Landing Rate of $0.685$. Adding AMP improves the raw action-space baseline from $\eta_{\mathrm{hit}}/ER=0.142/0.315$ to $0.285/0.465$, yet RoboStriker still wins 92.50\% of their cross-play bouts and reaches $0.685/0.824$, indicating that AMP alone does not account for the advantage of latent-space control. The underlying training dynamics, illustrated in Subfigure b of~\autoref{fig:training_convergence},
reveal the specific limitations of the baselines. The PPO-Only agent, lacking competitive co-evolution, exhibits passive behavior with a low Engagement Rate of $0.495$ and completely fails to discover proactive offensive strategies, yielding a landing rate of just $0.231$. Meanwhile, the Naive SP agent suffers from severe cyclical oscillations driven by non-transitive policy cycling and catastrophic forgetting, limiting its engagement to $0.580$. Although Fictitious SP mitigates these oscillations and improves the engagement to $0.650$, its average policy convergence still results in a significantly lower offensive landing rate of $0.420$ compared to our method. By integrating Neural Fictitious Self-Play within the latent manifold, RoboStriker successfully neutralizes policy cycling and smoothly converges to the highest tactical proficiency.

\subsection{Advantage of Latent-Space over Raw Action-Space}
\label{sec:latent_vs_action}
To isolate the contribution of latent-space abstraction from AMP regularization, we compare RoboStriker with direct 29-DoF action-space self-play with and without AMP. As shown in ~\autoref{fig:training_convergence}, the unregularized action-space baseline suffers from severe instability and poor offensive performance, while adding AMP substantially improves both physical stability and tactical metrics.

However, a clear gap remains between 29-DoF Action-Space + AMP and RoboStriker. Although AMP regularizes motion toward the reference distribution, strategic exploration is still performed directly in the high-dimensional joint space. In contrast, RoboStriker conducts exploration through the pretrained decoder over a structured latent manifold, achieving higher stability, engagement, striking efficiency, and torque smoothness. These results show that AMP and latent-space abstraction provide complementary benefits: AMP improves motion regularity, while the latent interface further structures the strategic search space.

\subsection{Real-World Deployment}
\label{sec:real_world}
\begin{figure}[t]
    \centering
    \includegraphics[width=0.85\linewidth]{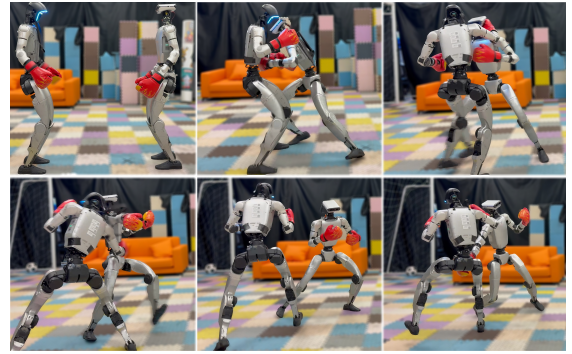}
    \caption{Real-world validation of RoboStriker.}
    \label{fig:real_world}
    \vspace{-4mm}
\end{figure}

Beyond simulation, we validate that the learned policies can transfer to physical hardware. We align simulation coefficients with measured physical properties and apply the domain randomization described above across all training stages. Following~\cite{liao2025beyondmimicmotiontrackingversatile}, joint PD gains are set proportional to inertia as $k_{p,j}=I_j\omega^{2}$ and $k_{d,j}=2I_j\zeta\omega$, with natural frequency $\omega=10$ and damping ratio $\zeta=2$. We deploy two Unitree G1 humanoids~\cite{unitreeg1} in a motion-capture environment: policies run at 50\,Hz via ONNX Runtime on the onboard CPUs and are coordinated through ROS\,2~\cite{macenski2022robot}. Pelvis markers tracked at 100\,Hz provide inter-robot spatial relations; combined with proprioception and forward kinematics, these yield the offensive target observations required by the boxing policy. To avoid abrupt torque discontinuities, both robots are initialized with a high-stiffness standing controller before transitioning to dynamic boxing. As illustrated in~\autoref{fig:real_world}, the robots execute stable, contact-rich maneuvers on hardware, confirming that the structured latent manifold isolates strategic reasoning from physical instability.

\begin{figure}[t]
    \centering
    \includegraphics[width=0.85\linewidth]{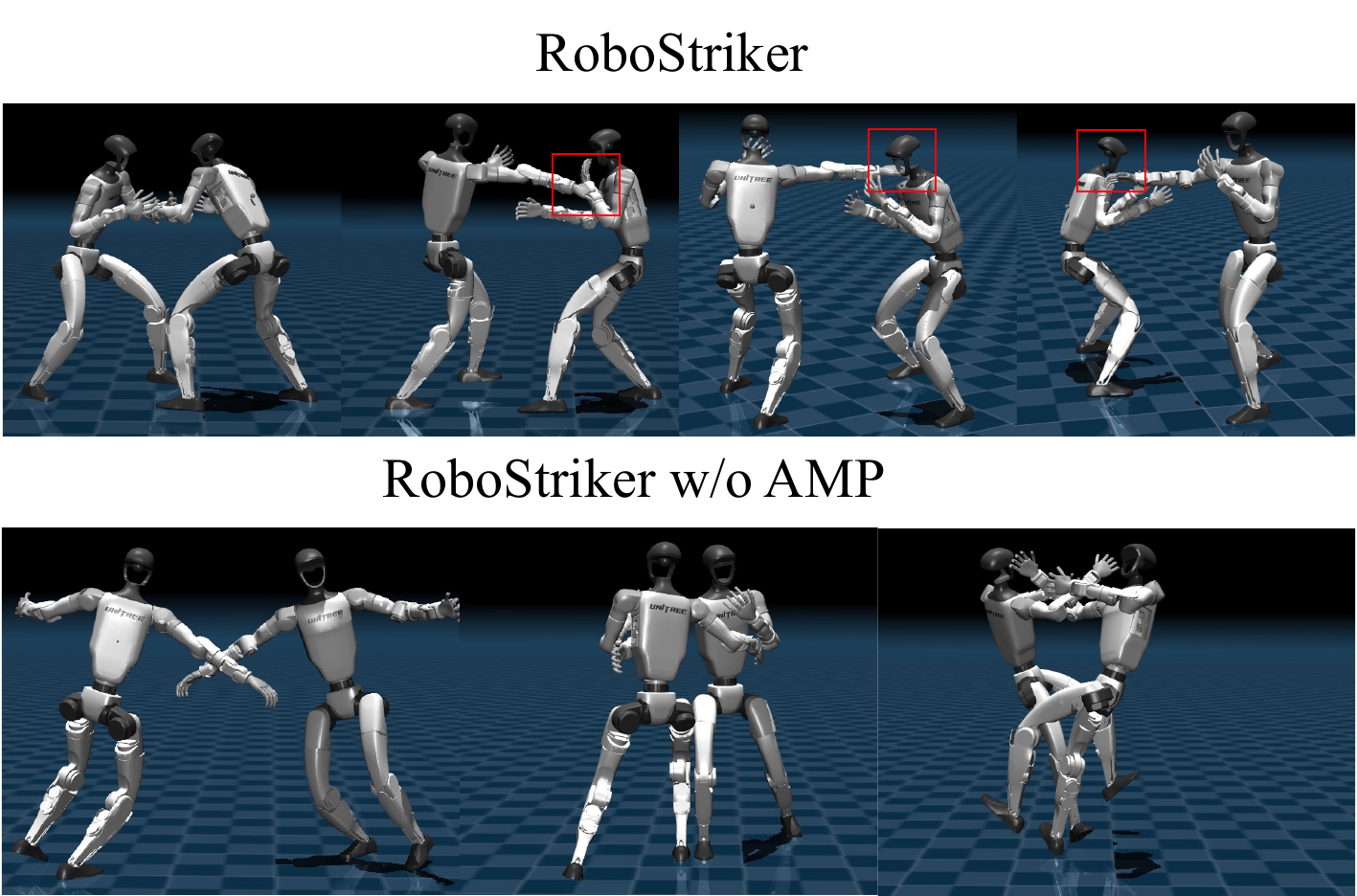}
    \caption{Mujoco visualization of RoboStriker and the one without AMP.}
    \label{fig:mujoco_visulization}
    \vspace{-4mm}
\end{figure}
\subsection{Ablation Study}
\label{sec:ablation_study}
We systematically evaluate our core components across~\autoref{tab:merged_results},~\autoref{fig:training_convergence}, and~\autoref{fig:mujoco_visulization}. First, as quantitatively recorded in~\autoref{tab:merged_results}, removing the AMP module severely degrades offensive efficiency despite maintaining high engagement. Furthermore, the corresponding training trajectories in~\autoref{fig:training_convergence} and simulation snapshots in~\autoref{fig:mujoco_visulization} reveal that without AMP, the agent aggressively approaches but executes erratic strikes, completely abandoning standard guards for physically unnatural postures.

Complementarily, adding AMP to the 29-DoF action-space baseline improves its stability and motion quality, but does not recover the performance of RoboStriker. Together, these ablations show that AMP regularization and
latent-space abstraction play complementary roles: AMP encourages human-like behavior, while the latent decoder provides a structured motor interface for strategic exploration.

Next, we validate the behavioral warm-up curriculum. As shown in~\autoref{fig:training_convergence}, initiating competitive self-play from scratch causes tactical performance to flatline near zero due to severe reward sparsity. Conversely, a Static-Target Specialist proficiently strikes passive targets but fails against dynamic opponents, yielding an overwhelming $95.38\%$ win rate for RoboStriker. Thus, the synergy of progressive curriculum design and AMP-guided movement is indispensable for producing a physically authentic and strategically dominant controller.
\begin{table}[t]
\centering
\caption{Ablation study on the latent dimension.}
\label{tab:latent_dim}
\begin{tabular}{lccc}
\toprule
\textbf{Latent Dim.} & \textbf{MPKPE (mm)} & \textbf{Win Rate vs.\ 32} & \bm{$\eta_{hit}$} \\
\midrule
8  & $128.2 \pm 4.4$ & $7.1\%$  & $0.328 \pm 0.03$ \\
16 & $85.9 \pm 1.7$  & $23.6\%$ & $0.334 \pm 0.03$ \\
\textbf{32 (Ours)} & $38.5 \pm 0.4$  & --      & $\mathbf{0.685 \pm 0.03}$ \\
64 & $\mathbf{35.0 \pm 0.4}$  & $39.2\%$ & $0.520 \pm 0.08$ \\
\bottomrule
\end{tabular}
\vspace{-4mm}
\end{table}
Finally, we evaluate the latent dimension trade-off between expressiveness and multi-agent stability. As illustrated in~\autoref{tab:latent_dim}, lower dimensions excessively constrain motion encoding, yielding rigid behaviors and high tracking discrepancies, quantified here by the Mean Per Keypoint Position Error(MPKPE). Expanding to 32 dimensions substantially improves kinematics, whereas 64 provides negligible physical gains. Strategically, this 64-dimensional expansion overcomplicates the continuous search manifold. The resulting representation redundancy slows convergence, degrading the offensive landing rate to 0.520. Ultimately, a 32-dimensional space offers the optimal balance, ensuring motion fidelity while preserving a compact manifold for stable strategic learning.

\section{Conclusion}

To address the conflict between multi-agent strategic exploration and humanoid physical fragility, we formulate competitive boxing as a decoder-induced latent zero-sum Markov game and provide an approximate-Nash analysis under standard regularity and approximate best-response assumptions. We instantiate this theoretical foundation through RoboStriker, a hierarchical framework that decouples strategic reasoning from low-level balance control via a topologically bounded manifold and drives competitive co-evolution using Latent-Space Neural Fictitious Self-Play. Extensive simulation evaluations, including cross-play tournaments and systematic ablations, demonstrate substantially improved physical stability and tactical proficiency over the evaluated baselines.

\printbibliography

@misc{schulman2017proximalpolicyoptimizationalgorithms,
      title={Proximal Policy Optimization Algorithms}, 
      author={John Schulman and Filip Wolski and Prafulla Dhariwal and Alec Radford and Oleg Klimov},
      year={2017},
      eprint={1707.06347},
      archivePrefix={arXiv},
      primaryClass={cs.LG},
      url={https://arxiv.org/abs/1707.06347}, 
}

@misc{ross2011reductionimitationlearningstructured,
      title={A Reduction of Imitation Learning and Structured Prediction to No-Regret Online Learning}, 
      author={Stephane Ross and Geoffrey J. Gordon and J. Andrew Bagnell},
      year={2011},
      eprint={1011.0686},
      archivePrefix={arXiv},
      primaryClass={cs.LG},
      url={https://arxiv.org/abs/1011.0686}, 
}

@misc{unitreeg1,
  title = {Unitree g1 humanoid agent ai avatar},
  author = {Unitree},
  url = "https://www.unitree.com/g1",
  year = 2024,
}

@article{chen2025gmt,
title={GMT: General Motion Tracking for Humanoid Whole-Body Control},
author={Chen, Zixuan and Ji, Mazeyu and Cheng, Xuxin and Peng, Xuanbin and Peng, Xue Bin and Wang, Xiaolong},
journal={arXiv:2506.14770},
year={2025}
}

@misc{liao2025beyondmimicmotiontrackingversatile,
      title={BeyondMimic: From Motion Tracking to Versatile Humanoid Control via Guided Diffusion}, 
      author={Qiayuan Liao and Takara E. Truong and Xiaoyu Huang and Guy Tevet and Koushil Sreenath and C. Karen Liu},
      year={2025},
      eprint={2508.08241},
      archivePrefix={arXiv},
      primaryClass={cs.RO},
      url={https://arxiv.org/abs/2508.08241}, 
}

@misc{xsens_mvn,
  title  = {Xsens MVN: Inertial motion capture system},
  author = {{Xsens Technologies}},
  year   = {2024},
  note   = {Accessed: 2026-01-28},
  url    = {https://www.xsens.com/motion-capture}
}

@article{samuel1959some,
  title={Some studies in machine learning using the game of checkers},
  author={Samuel, Arthur L},
  journal={IBM Journal of research and development},
  volume={3},
  number={3},
  pages={210--229},
  year={1959},
  publisher={IBM}
}

@inproceedings{hernandez2019generalized,
  title={A generalized framework for self-play training},
  author={Hernandez, Daniel and Denamgana{\"\i}, Kevin and Gao, Yuan and York, Peter and Devlin, Sam and Samothrakis, Spyridon and Walker, James Alfred},
  booktitle={2019 IEEE Conference on Games (CoG)},
  pages={1--8},
  year={2019},
  organization={IEEE}
}

@article{peng2018deepmimic,
  title={Deepmimic: Example-guided deep reinforcement learning of physics-based character skills},
  author={Peng, Xue Bin and Abbeel, Pieter and Levine, Sergey and Van de Panne, Michiel},
  journal={ACM Transactions On Graphics (TOG)},
  volume={37},
  number={4},
  pages={1--14},
  year={2018},
  publisher={ACM New York, NY, USA}
}

@article{peng2021amp,
  title={Amp: Adversarial motion priors for stylized physics-based character control},
  author={Peng, Xue Bin and Ma, Ze and Abbeel, Pieter and Levine, Sergey and Kanazawa, Angjoo},
  journal={ACM Transactions on Graphics (ToG)},
  volume={40},
  number={4},
  pages={1--20},
  year={2021},
  publisher={ACM New York, NY, USA}
}

@article{heinrich2016deep,
  title={Deep reinforcement learning from self-play in imperfect-information games},
  author={Heinrich, Johannes and Silver, David},
  journal={arXiv preprint arXiv:1603.01121},
  year={2016}
}

@article{luo2023universal,
  title={Universal humanoid motion representations for physics-based control},
  author={Luo, Zhengyi and Cao, Jinkun and Merel, Josh and Winkler, Alexander and Huang, Jing and Kitani, Kris and Xu, Weipeng},
  journal={arXiv preprint arXiv:2310.04582},
  year={2023}
}

@article{peng2022ase,
  title={Ase: Large-scale reusable adversarial skill embeddings for physically simulated characters},
  author={Peng, Xue Bin and Guo, Yunrong and Halper, Lina and Levine, Sergey and Fidler, Sanja},
  journal={ACM Transactions On Graphics (TOG)},
  volume={41},
  number={4},
  pages={1--17},
  year={2022},
  publisher={ACM New York, NY, USA}
}

@article{brown1951iterative,
  title={Iterative solution of games by fictitious play},
  author={Brown, George W},
  journal={Act. Anal. Prod Allocation},
  volume={13},
  number={1},
  pages={374},
  year={1951}
}

@article{yin2025unitracker,
  title={Unitracker: Learning universal whole-body motion tracker for humanoid robots},
  author={Yin, Kangning and Zeng, Weishuai and Fan, Ke and Dai, Minyue and Wang, Zirui and Zhang, Qiang and Tian, Zheng and Wang, Jingbo and Pang, Jiangmiao and Zhang, Weinan},
  journal={arXiv preprint arXiv:2507.07356},
  year={2025}
}

@article{zhang2025track,
  title={Track any motions under any disturbances},
  author={Zhang, Zhikai and Guo, Jun and Chen, Chao and Wang, Jilong and Lin, Chenghuai and Lian, Yunrui and Xue, Han and Wang, Zhenrong and Liu, Maoqi and Lyu, Jiangran and others},
  journal={arXiv preprint arXiv:2509.13833},
  year={2025}
}

@inproceedings{tessler2023calm,
  title={Calm: Conditional adversarial latent models for directable virtual characters},
  author={Tessler, Chen and Kasten, Yoni and Guo, Yunrong and Mannor, Shie and Chechik, Gal and Peng, Xue Bin},
  booktitle={ACM SIGGRAPH 2023 Conference Proceedings},
  pages={1--9},
  year={2023}
}

@misc{nvidia2025isaaclabgpuacceleratedsimulation,
      title={Isaac Lab: A GPU-Accelerated Simulation Framework for Multi-Modal Robot Learning}, 
      author={NVIDIA and : and Mayank Mittal and Pascal Roth and James Tigue and Antoine Richard and Octi Zhang and Peter Du and Antonio Serrano-Muñoz and Xinjie Yao and René Zurbrügg and Nikita Rudin and Lukasz Wawrzyniak and Milad Rakhsha and Alain Denzler and Eric Heiden and Ales Borovicka and Ossama Ahmed and Iretiayo Akinola and Abrar Anwar and Mark T. Carlson and Ji Yuan Feng and Animesh Garg and Renato Gasoto and Lionel Gulich and Yijie Guo and M. Gussert and Alex Hansen and Mihir Kulkarni and Chenran Li and Wei Liu and Viktor Makoviychuk and Grzegorz Malczyk and Hammad Mazhar and Masoud Moghani and Adithyavairavan Murali and Michael Noseworthy and Alexander Poddubny and Nathan Ratliff and Welf Rehberg and Clemens Schwarke and Ritvik Singh and James Latham Smith and Bingjie Tang and Ruchik Thaker and Matthew Trepte and Karl Van Wyk and Fangzhou Yu and Alex Millane and Vikram Ramasamy and Remo Steiner and Sangeeta Subramanian and Clemens Volk and CY Chen and Neel Jawale and Ashwin Varghese Kuruttukulam and Michael A. Lin and Ajay Mandlekar and Karsten Patzwaldt and John Welsh and Huihua Zhao and Fatima Anes and Jean-Francois Lafleche and Nicolas Moënne-Loccoz and Soowan Park and Rob Stepinski and Dirk Van Gelder and Chris Amevor and Jan Carius and Jumyung Chang and Anka He Chen and Pablo de Heras Ciechomski and Gilles Daviet and Mohammad Mohajerani and Julia von Muralt and Viktor Reutskyy and Michael Sauter and Simon Schirm and Eric L. Shi and Pierre Terdiman and Kenny Vilella and Tobias Widmer and Gordon Yeoman and Tiffany Chen and Sergey Grizan and Cathy Li and Lotus Li and Connor Smith and Rafael Wiltz and Kostas Alexis and Yan Chang and David Chu and Linxi "Jim" Fan and Farbod Farshidian and Ankur Handa and Spencer Huang and Marco Hutter and Yashraj Narang and Soha Pouya and Shiwei Sheng and Yuke Zhu and Miles Macklin and Adam Moravanszky and Philipp Reist and Yunrong Guo and David Hoeller and Gavriel State},
      year={2025},
      eprint={2511.04831},
      archivePrefix={arXiv},
      primaryClass={cs.RO},
      url={https://arxiv.org/abs/2511.04831}, 
}

@inproceedings{hershey2007approximating,
  title={Approximating the Kullback Leibler divergence between Gaussian mixture models},
  author={Hershey, John R and Olsen, Peder A},
  booktitle={2007 IEEE International Conference on Acoustics, Speech and Signal Processing-ICASSP'07},
  volume={4},
  pages={IV--317},
  year={2007},
  organization={IEEE}
}

@misc{araujo2025retargetingmattersgeneralmotion,
      title={Retargeting Matters: General Motion Retargeting for Humanoid Motion Tracking}, 
      author={Joao Pedro Araujo and Yanjie Ze and Pei Xu and Jiajun Wu and C. Karen Liu},
      year={2025},
      eprint={2510.02252},
      archivePrefix={arXiv},
      primaryClass={cs.RO},
      url={https://arxiv.org/abs/2510.02252}, 
}

@misc{zeng2025behaviorfoundationmodelhumanoid,
      title={Behavior Foundation Model for Humanoid Robots}, 
      author={Weishuai Zeng and Shunlin Lu and Kangning Yin and Xiaojie Niu and Minyue Dai and Jingbo Wang and Jiangmiao Pang},
      year={2025},
      eprint={2509.13780},
      archivePrefix={arXiv},
      primaryClass={cs.RO},
      url={https://arxiv.org/abs/2509.13780}, 
}

@inproceedings{macenski2022robot,
  title     = {Robot Operating System 2: Design, Architecture, and Uses in the Wild},
  author    = {Macenski, Steve and Foote, Tully and Gerkey, Brian and Lalancette, C{\'e}dric and Woodall, William},
  booktitle = {Science Robotics},
  volume    = {7},
  number    = {66},
  year      = {2022}
}

@article{10.1145/3450626.3459761,
author = {Won, Jungdam and Gopinath, Deepak and Hodgins, Jessica},
title = {Control strategies for physically simulated characters performing two-player competitive sports},
year = {2021},
issue_date = {August 2021},
publisher = {Association for Computing Machinery},
address = {New York, NY, USA},
volume = {40},
number = {4},
issn = {0730-0301},
url = {https://doi.org/10.1145/3450626.3459761},
doi = {10.1145/3450626.3459761},
journal = {ACM Trans. Graph.},
month = jul,
articleno = {146},
numpages = {11}
}

@misc{zhang2026learningathletichumanoidtennis,
      title={Learning Athletic Humanoid Tennis Skills from Imperfect Human Motion Data}, 
      author={Zhikai Zhang and Haofei Lu and Yunrui Lian and Ziqing Chen and Yun Liu and Chenghuai Lin and Han Xue and Zicheng Zeng and Zekun Qi and Shaolin Zheng and Qing Luan and Jingbo Wang and Junliang Xing and He Wang and Li Yi},
      year={2026},
      eprint={2603.12686},
      archivePrefix={arXiv},
      primaryClass={cs.RO},
      url={https://arxiv.org/abs/2603.12686}, 
}
\end{document}